\documentclass{article}

\usepackage{PRIMEarxiv}

\usepackage[utf8]{inputenc} 
\usepackage[T1]{fontenc}    
\usepackage{hyperref}       
\usepackage{url}            
\usepackage{booktabs}       
\usepackage{amsfonts}       
\usepackage{nicefrac}       
\usepackage{microtype}      
\usepackage{lipsum}
\usepackage{fancyhdr}       
\usepackage{graphicx}       
\graphicspath{{media/}}     
\usepackage{amsmath}        

\title{Schema-R1: A reasoning training approach for schema linking in Text-to-SQL Task
}

\author{
  Wuzhenghong Wen, Su Pan, Yuwei Sun \\
  School of Internet of Things, Nanjing University of Posts and Telecommunications \\
  Nanjing \\
  China\\
  \texttt{\{Wuzhenghong Wen, Su Pan, yuwei Sun\}2022070804@njupt.edu.cn} \\
}

\begin{document}
\maketitle

\begin{abstract}
Schema linking is a critical step in Text-to-SQL task, aiming to accurately predict the table names and column names required for the SQL query based on the given question. However, current fine-tuning approaches for schema linking models employ a rote-learning paradigm, excessively optimizing for ground truth schema linking outcomes while compromising reasoning ability. This limitation arises because of the difficulty in acquiring a high-quality reasoning sample for downstream tasks. To address this, we propose Schema-R1, a reasoning schema linking model trained using reinforcement learning.
Specifically, Schema-R1 consists of three key steps: constructing small batches of high-quality reasoning samples, supervised fine-tuning for cold-start initialization, and rule-based reinforcement learning training.
The final results demonstrate that our method effectively enhances the reasoning ability of the schema linking model, achieving a 10\% improvement in filter accuracy compared to the existing method. Our code is available at https://github.com/hongWin/Schema-R1/.
\end{abstract}

\keywords{Text-to-SQL \and Reasoning model \and Fine-Tuning}

\section{Introduction}

Text-to-SQL is a Natural Language Processing (NLP) task that converts natural language questions into structured query language (SQL). This process explicitly or implicitly involves: schema linking, SQL generation, coding enhancement, and self-correction~\cite{shi2024survey,ma2025sql}.
Schema linking is a key step in Text-to-SQL, where the system filters the database tables and columns most relevant to the user's natural language question to support accurate SQL generation~\cite{lei-etal-2020-examining}.

According to existing research, to achieve the goal of Text-to-SQL, the schema linking task can be implicitly incorporated into the Text-to-SQL reasoning process~\cite{DAIL-SQL,xue2023db,chen2024open,meta}, or can be explicitly constructed using an additional schema linking generation model~\cite{DIN-SQL,lee-etal-2025-mcs,zhang2023act,pourreza2024dts}.However, while implicitly integrating schema linking into SQL generation ensures reasoning coherence, it complicates the construction of effective supervision signals for erroneous schema linking, making corrections difficult. Thus, separating schema linking from SQL generation and adopting a pipeline approach enhances interpretability and facilitates error correction in text-to-SQL tasks.~\cite{DIN-SQL,lee-etal-2025-mcs,zhang2023act} employ prompt engineering techniques to utilize commercial large language model(LLM) as schema linking models to predict potential database tables and columns. Although these commercial models benefit from extensive knowledge bases and robust training data, their application often encounters significant limitations when dealing with large-scale databases that feature complex structures and stringent privacy requirements. Therefore, to ensure data privacy while maintaining strong adaptability, open source LLMs are increasingly becoming key collaborators in vertical domains~\cite{sun2023instruction}. Supervised fine-tuning (SFT) has emerged as the primary method to equip these open-source models with specialized domain knowledge.~\cite{gorti2024msc,pourreza2024dts,wu2024datagpt,li2024codes,CogSQL} employ supervised fine-tuning (SFT) to train specialized schema-linking models on open-source LLMs for text-to-SQL tasks. The training pipeline is illustrated in Fig.~\ref{fig:pic1}:
\begin{figure}[t!]
    \centering
    \includegraphics[width=1\linewidth]{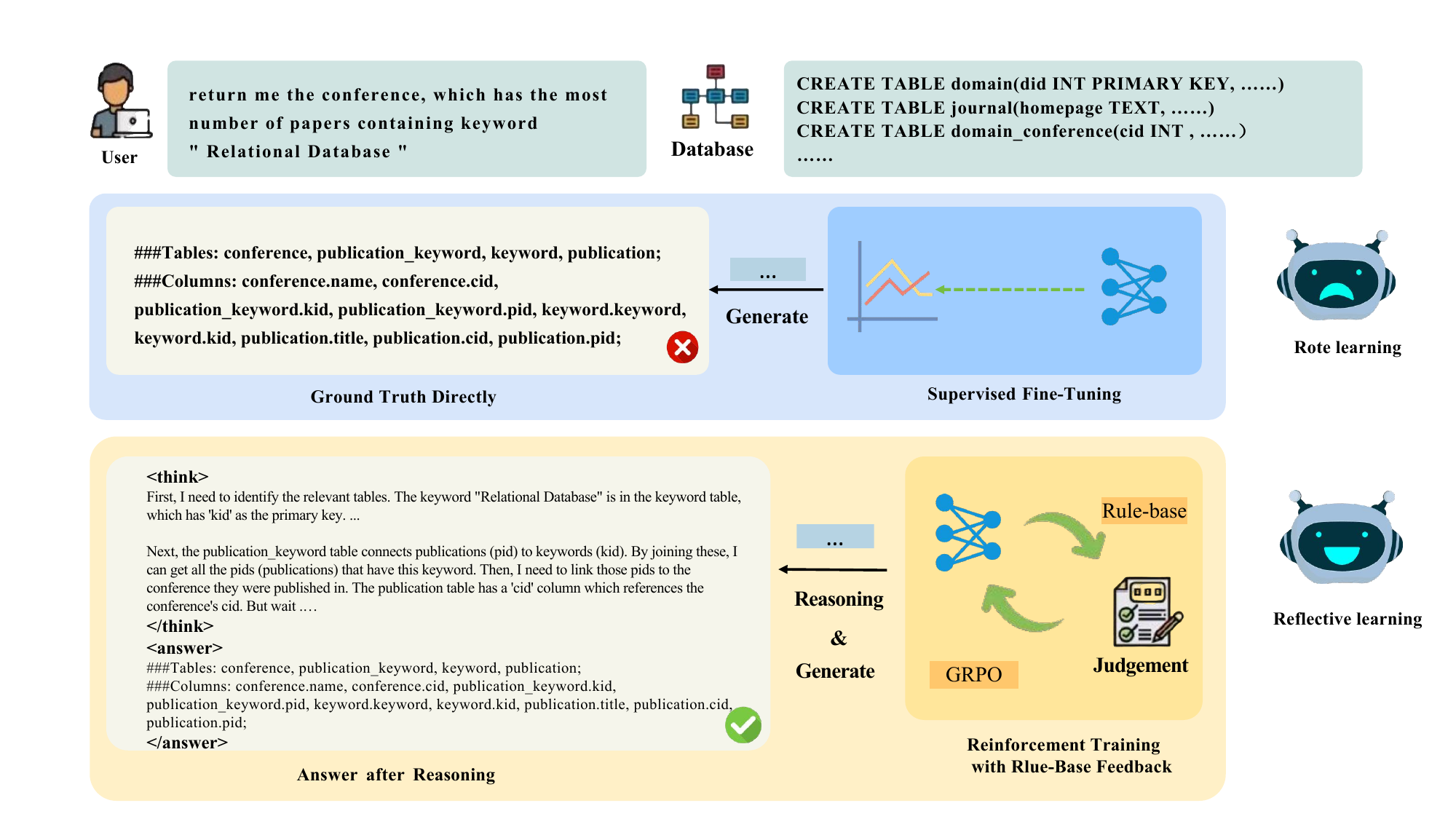} 
    \caption{The difference between SFT and Reasoning RL in Schema linking downstream task tuning}
    \label{fig:pic1} 
\end{figure}
As shown in Fig.~\ref{fig:pic1}, directly employing supervised fine-tuning (SFT) to build goal-oriented schema-linking models results in a textbook-style training approach. This causes the model to overfit to the supervised objectives while losing its inherent reasoning capabilities during inference. However, constructing chain-of-thought (CoT)~\cite{wei2022chain} based supervision targets for fine-tuning is constrained by the limited availability of high-quality CoT samples~\cite{guo2025deepseek}.

Reinforcement Learning(RL) for reasoning training is demonstrated remarkable capabilities in self-generated reasoning and robust adaptation to downstream tasks across various LLM applications. As implemented in Reasoning Reinforcement Learning~\cite{shao2024deepseekmath,hu2025reinforce++,yu2025dapo}, this approach generates multiple reasoning paths and employs group optimization strategies to reinforce high-reward trajectories while suppressing low-reward ones. Compared to supervised fine-tuning (SFT) that optimizes for a single objective, RL for reasoning training exhibits superior robustness in downstream task training.

To develop a schema linking model with robust reasoning capabilities, we propose Schema-R1, a novel approach that implements a three-stage training framework:
\textbf{\textit{Prompt-based Knowledge Enhancement:}} We first design a set of prompt templates to leverage commercial LLMs to generate coherent reasoning information based on ground truth schema linking information, effectively compensating for missing reasoning information.
\textbf{\textit{Supervised Fine-tuning for cold start:}} A small subset of dataset samples is then carefully selected to construct SFT tasks, where each training instance contains the question, database schema, reasoning information, and ground-truth schema linking information, enabling the model to learn proper output patterns.
\textbf{\textit{Reinforcement Learning(RL) for reasoning training:}} We employ GPRO with the SFTed model as reference, dynamically capturing the table and column prediction accuracy during training to establish the corresponding reward signals. This guides the model to progressively favor higher-reward reasoning paths while refining its reasoning capabilities.
Experimental results in Spider-dev demonstrate that our approach achieves at least 10\% improvement in both table and column filter accuracy compared to the state-of-the-art schema linking method based on fine-tuning.In summary, our contributions can be characterized as follows. 

\textbf{\textit{Weak Reasoning Model Initialization:}} Through prompt engineering on limited samples, we use commercial LLM combined with supervised fine-tuning (SFT) to develop a preliminary schema linking model with basic reasoning capabilities that strictly adheres to instructional constraints.

\textbf{\textit{Schema-R1: Advanced Reasoning Model:}}
We propose Schema-R1, a robust reasoning-enhanced schema linking model that employs cold start fine-tuning with minimal samples and self-optimizes through our novel reasoning RL framework for CoT refinement.

\section{Related Work}  

\textbf{\textit{Scheme linking model}} Early research on schema linking models primarily focused on developing encoder-decoder-based architectures. For example,~\cite{RESDSQL} employed an encoder to encode both the question and the target database, with the aim of training the model to capture the relevance of tables and columns in different questions. Meanwhile,~\cite{fu2023catsql} utilized multiple decoders to simultaneously address various subtasks of Text-to-SQL. 
In contrast to these approaches,~\cite{wang-etal-2020-rat,LGESQL} encode schema graphs to represent the connection information within the databases and then use a decoder to infer potential schema information. With the growing demonstration of powerful reasoning and knowledge capabilities in both commercial and open-source LLMs, an increasing number of studies are leveraging prompt engineering to enhance the reasoning abilities of these models and apply them to schema linking tasks. ~\cite{dong2023c3} explored GPT-4's performance as a schema linking model using a zero-shot approach, while ~\cite{DIN-SQL} adopted an N-shot method. ~\cite{lee-etal-2025-mcs,tan-etal-2024-enhancing} employed multiple prompts to generate schema linking information and then aggregated multiple reasoning paths to derive the final answer. Similarly, ~\cite{zhang2023act} utilize self-generated CoT reasoning to improve the probability of obtaining optimal schema linking results. In contrast to these methods, ~\cite{li2024pet,Schema-Enhances} first generated SQL queries and then inferred the schema linking information in reverse.

Although commercial LLMs demonstrate superior performance on various open source tasks, they often require supervised fine-tuning (SFT) using open-source models when applied to private databases or highly specialized domains.Several research utilize SFT to map natural language questions and database schemas to target tables and columns. ~\cite{gorti2024msc,pourreza2024dts,wu2024datagpt,li2024codes,CogSQL} developed end-to-end supervised fine-tuning approaches for this purpose, while ~\cite{shen2023spsql} introduced a two-stage pipeline that first identifies relevant tables before predicting columns. ~\cite{zhang2024tablellm} take a more direct approach by formulating tables prediction tasks as fine-tuning targets.However, current supervised fine-tuning methods for schema linking often suffer from a critical limitation: the lack of high-quality reasoning information in training data frequently reduces the process to rote memorization rather than genuine schema understanding.

\textbf{\textit{RL for reasoning}} Research on reasoning in LLM has evolved significantly, progressing from linear prompt-based causal enhancement~\cite{wei2022chain} to heuristic reasoning search methods~\cite{yao2023tree,jin-etal-2024-graph}, and more recently to the trending approach of CoT prompting combined with RL training ~\cite{shao2024deepseekmath}.GRPO ~\cite{shao2024deepseekmath} represents a notable advancement in this direction. To address training instability issues in GRPO, subsequent work ~\cite{hu2025reinforce++,yu2025dapo} proposes various improvements and optimizations. In addition, researchers ~\cite{fang2025thinkless,jiang2025think,ma2025reasoning} also explore injecting reasoning rules into models through RL training.
The application of RL for reasoning demonstrate particularly strong momentum in specialized domains, including: Mathematics ~\cite{acemath2024,wen2025light}, Rewards model~\cite{chen2025rm,liu2025inference}, Financial applications ~\cite{liu2025fin}. Meanwhile,~\cite{yang2025table,lei2025reasoning} successfully applied RL-based reasoning approaches to table prediction tasks in the Text-to-SQL task. Unlike \cite{yang2025table,lei2025reasoning}, Schema R1 shows improved capabilities by performing not only table prediction, but also column prediction.

\section{Preliminaries}
\subsection{supervise fine-tuning in schema linking task}
Following \cite{pourreza2024dts}'s approach to supervised fine-tuning task design, the schema linking task's fine tuning objective can be defined as:
\begin{equation}  
\setlength{\abovedisplayskip}{3pt}
\setlength{\belowdisplayskip}{3pt}
	\begin{aligned}  
		& \underset{\theta}{\text{minimize}}  
		& & \frac{1}{N} \sum_{i=1}^{N} \text{Loss}((t^{*}_i,c^{*}_i), \text{LLM}_S(q_i, S_i; \theta))  
	\end{aligned}  
    \label{schema_link_0}
\end{equation}
Eq.~\eqref{schema_link_0} enable the schema linking LLM $LLM_S$ to predict tables ${t^{*}}_{i}$ and columns ${c^{*}}_{i}$ that match the SQL query requirements $q_{i}$, given total database descriptions $S_i$ in the database. $\theta$ represents all the parameters involved in the training and $Loss$ is the cross-entropy loss.

\subsection{RL For Reasoning}
As one of the most influential reasoning RL algorithms, GRPO effectively leverages the exploration-exploitation trade-off inherent in reinforcement learning.
To optimize GPU efficiency, it eliminates the requirement for a value model to evaluate the advantage for an individual sampled trajectories. Instead, it computes rewards for a group of sampled trajectories under the same problem instance, then derives their relative advantages from these rewards. The optimization objective of GRPO can be formally expressed as:
\begin{equation}  
\begin{aligned} \mathcal{J}_{G R P O}(\theta)= & \mathbb{E}\left[q \sim P(Q),\left\{o_i\right\}_{i=1}^G \sim \pi_{\theta_{\text {old}}}(O \mid q)\right] \\ & \frac{1}{G} \sum_{i=1}^G \frac{1}{\left|o_i\right|} \sum_{t=1}^{\left|o_i\right|}\left\{\min \left[\frac{\pi_\theta\left(o_{i, t} \mid q, o_{i,<t}\right)}{\pi_{\theta_{\text {old }}}\left(o_{i, t} \mid q, o_{i,<t}\right)} \hat{A}_{i, t,} \operatorname{clip}\left(\frac{\pi_\theta\left(o_{i, t} \mid q, o_{i,<t}\right)}{\pi_{\theta_{\text {old }}}\left(o_{i, t} \mid q, o_{i,<t}\right)}, 1-\varepsilon, 1+\varepsilon\right) \hat{A}_{i, t}\right]-\beta \mathbb{D}_{K L}\left[\pi_\theta| | \pi_{r e f} \right]\right\},\end{aligned}
\label{grpo}
\end{equation}


As shown in Eq.~\eqref{grpo}, $q \sim P(Q)$: Denotes the input question sampled from the question distribution $\mathrm{P}(\mathrm{Q})$. 
$G$ is the group size, which represents the number of different candidate responses generated for the same question. 
$\pi_\theta$ denotes the current policy model,whose parameters $\theta$ are updated by optimizing the objective function, while $\pi_{\theta_{\text{old}}}$ denotes the old policy model, with fixed parameters, used to generate training samples.
$\hat{A}_{i, t}$ is the relative advantage value within the group, computed by normalizing the rewards, which is the most significant difference compared to PPO\cite{PPO} .

\section{Proposed Method}
\subsection{Preparation For Cold Start}
\label{subsec:prepare_sft}
In our study, to enhance GRPO's answer generation alignment with human preferences and accelerate convergence, we employe supervised fine-tuning to ensure the model adheres to specific output formats. However, since traditional schema linking models typically rely on rote memorization of answers, we develop a reasoning-oriented fine-tuning approach.

To achieve this, we first curated high-quality reasoning information by extracting 200 samples from the Spider dataset. These samples were structured into prompt templates that contain the question, database schema, and target tables / columns. The objective is to guide commercial LLMs (specifically, we selected DeepSeek-R1) to reconstruct the missing reasoning process using ground truth schema linking as supervision, The detailed construction of prompt templates is provided in the Appendix.~\ref{app:slice-construction}.
Building upon the aforementioned mini-dataset, we formulate our supervised fine-tuning objective as presented in Eq.~\eqref{schema_link_1}.
\begin{equation}  
\setlength{\abovedisplayskip}{3pt}
\setlength{\belowdisplayskip}{3pt}
	\begin{aligned}  
		& \underset{\theta}{\text{minimize}}  
		& & \frac{1}{N} \sum_{i=1}^{N} \text{Loss}((t^{*}_i,c^{*}_i,CoT_i), \text{LLM}_S(q, S, D; \theta))  
	\end{aligned}  
    \label{schema_link_1}
\end{equation}
As specified in Eq.~\eqref{schema_link_1}, $CoT_i$ represents the ground truth CoT corresponding to the given question. Through our supervised fine-tuning approach, the model generates responses that reliably adhere to our desired format: encapsulating reasoning processes within <think>...</think> tags and schema linking outputs within <answer>...</answer> tags. This structured output enables the subsequent reasoning RL phase to precisely extract supervisory signals from the <answer>...</answer> segments while simultaneously training the reasoning capabilities demonstrated in the <think>...</think> components.

\subsection{Rule Base For Reasoning RL}
The design of appropriate reward objectives is crucial to guide the final learning outcomes of RL models. As GRPO is a result-supervised RL algorithm, we establish rule-based scoring for GRPO's rewards based on the output within the <answer>...</answer> tags. In Schema-R1 training, we implement a comprehensive reward system consisting of three distinct components.

\textbf{\textit{Format Reward}} 
Although the model acquires basic instruction following capabilities through cold start initialization, RL training tends to stimulate its exploratory behavior during the training process. This encourages more creative responses but may simultaneously cause the model to deviate from our desired output format, leading to response mismatches. To maintain a balance between creative exploration and format compliance, we implement a rule-based mechanism that continuously reinforces the target response paradigm. The specific configuration of our reward is as follows:
\begin{equation}
R_f = 
\begin{cases} 
1, & \text{format success} \\
0, & \text{format fail} \\
\end{cases}
\label{S_f}
\end{equation}

\begin{equation}
\begin{aligned}
R_c = bool(count\text{("\#\#\#table:") == 1}) + bool(count\text{("\#\#\#columns:") == 1})
\end{aligned}
\label{S_e}
\end{equation}

The format reward comprises two components: Eq.~\eqref{S_f} ensures that the sampled response correctly encloses the answer in <answer>...</answer> tags and the reasoning in <think>...</think> tags, while Eq.~\eqref{S_e} directs the model to place table predictions after text{"\#\#\#table:"} and column predictions after text{"\#\#\#column:"}, with $count$ representing a string frequency function.

\textbf{\textit{Reasoning Length Reward}} Overly lengthy reasoning process increases inference and training time but may not enhance performance ~\cite{aggarwal2025l1, hou2025thinkprune}, while insufficient reasoning hinders complete problem solving. Thus, we set an optimal reasoning length baseline, constructed as follows:
\begin{equation}
R_l = 
\begin{cases} 
0, & len_{response} < \text{Lower Length } \\
1, & \text{Lower Length } <= len_{response} < \text{Upper Length} \\
0, &  \text{Upper Length } <= len_{response}  \\
\end{cases}
\label{S_L}
\end{equation}

As shown in Eq.~\eqref{S_L}, $len_{response}$ denotes the total token count of the response, while Lower Length and Upper Length represent the lower and upper bounds of the desired response length, respectively. 

\textbf{\textit{Schema Linking Reward}}
For the schema linking task, where the objective is to determine whether the predicted tables and columns accurately match the ground truth, we adopt the evaluation approach of \cite{pourreza2024dts}. Specifically, we utilize the filter accuracy metric (filter acc) to configure task-specific rewards and penalties.
\begin{equation}
\begin{aligned}
R_{st} = R_{tmax}/len(t^{*}_{i}) * len(t^{*}_{i} - t^{'}_{i}) - 
P_{tmax}/len(t^{'}_{i}) * len(t^{'}_{i} - t^{*}_{i})
\end{aligned}
\label{S_t}
\end{equation}

\begin{equation}
\begin{aligned}
R_{sc} = R_{cmax}/len(c^{*}_{i}) * len(c^{*}_{i} - c^{'}_{i}) - 
P_{cmax}/len(c^{'}_{i}) * len(c^{'}_{i} - c^{*}_{i})
\end{aligned}
\label{S_c}
\end{equation}

In Eq.~\eqref{S_t}, $t^{'}_{i}$ denotes the predicted set of tables, while $t^{*}_{i}$ represents the ground truth set of tables. From these we can derive the correctly predicted table set ($t^{*}_{i} - t^{'}_{i}$) and the incorrectly predicted set ($t^{'}_{i} - t^{*}_{i}$).
We define a maximum achievable reward $R_{tmax}$ and a maximum penalty $P_{tmax}$ for the table predictions. Consequently, the average reward for each correctly predicted item is calculated as $P_{tmax}/len(t^{*}_{i})$, and the average penalty for each incorrect prediction is $P_{tmax}/len(t^{'}_{i})$. The final reward for the prediction of the table $R_{st}$, is then given by the difference between the total rewards $R_{tmax}/len(t^{*}_{i}) * len(t^{*}_{i} - t^{'}_{i})$ and the penalties $P_{tmax}/len(t^{'}_{i}) * len(t^{'}_{i} - t^{*}_{i})$.
Similarly, the reward for column prediction $R_{sc}$ can be formulated in the same manner as in Eq.~\eqref{S_c}.
In the schema linking task, accurate table prediction is crucial for ensuring correct column predictions in subsequent steps. To reflect this importance, we set $R_{tmax} > R_{cmax}$ and $P_{tmax} > P_{cmax}$. The overall reward for schema linking is thus expressed as: $R_{s}  = R_{st} + R_{sc}$.

\section{Experiment}
\subsection{Setup}
\textbf{\textit{Benchmarks}} 
We conducted our training using Schema-R1 in the Spider dataset. Following the methodology outlined in ~\cite{pourreza2024dts}, we constructed the schema linking task by extracting ground truth tables and columns from the SQL queries. After validation, we compile a total of 8529 training samples, with the first 200 samples reserved for creating CoT supervised fine-tuning data and the remaining 8329 samples used for GRPO training. For validation, we applied the same approach to extract ground truth tables and columns from the Spider-dev set. 
The evaluation metrics we used are the following.
Exact Match (EM): measures whether the predicted content fully aligns with the ground truth.

Filtered Accuracy (FilteredAcc): quantifies the overlap between predictions and ground truth, with higher values indicating greater coverage of the correct content.

Recall (Rec): assess the quality of the prediction of the model in terms of completeness.

\textbf{\textit{training setting}}
Our experiments were based on the Qwen2.5-0.5B and Qwen2.5-1.5B foundation models. During the cold-start SFT, we employed full-parameter fine-tuning with a conservative learning rate of 5e-5 and limited the training to 3 epochs to mitigate overfitting. For the GRPO training stage, we adopted a sampling strategy that generated 10 responses per query, maintained the learning rate at 2e-05, set the training batch size to 10. The 0.5B model was trained on three A100-40G GPUs, whereas the 1.5B model utilized three A100-80G GPUs. VLLM was used to speed up inference, with a temperature setting of 1.

\subsection{Main Results}
\begin{table*}[h!]  
\renewcommand{\tabcolsep}{20pt}
	\centering  
    \renewcommand{\arraystretch}{0.75}
	\caption{Performance of table prediction for different models in the Table prediction Task.} 
\resizebox{0.99\textwidth}{!}{
	\begin{tabular}{l|ccc}  
		\toprule  
		\multicolumn{4}{c}{\textbf{Table Prediction}} \\  
		\midrule  
		Method & \centering EM & Filtered \newline Acc &  Rec \\  
		\midrule  
		Qwen2.5-0.5B (DTS-SQL) & 54.28 & 64.24 &  74.07  \\  
		Qwen2.5-1.5B (DTS-SQL) & 64.84 & 75.0 & 83.03  \\  \midrule
		Qwen2.5-0.5B (Cold-start) & 29.98 & 53.78 &  65.02  \\    
		Qwen2.5-1.5B (Cold-start)  & 56.67 & 70.41 &  79.06  \\  \midrule 
		Qwen2.5-0.5B (Schema-R1) & 55.38 & 75.60 &  85.34  \\  
		Qwen2.5-1.5B (Schema-R1) & \textbf{73.21} &  \textbf{89.94} & \textbf{94.40}  \\  
		\bottomrule  
	\end{tabular} }
    \label{box1}
    \vspace{-12pt}
\end{table*} 

\begin{table*}[h!]  
\renewcommand{\tabcolsep}{20pt}
	\centering  
    \renewcommand{\arraystretch}{0.75}
	\caption{Performance of table prediction for different method in Columns prediction Task.} 
\resizebox{0.99\textwidth}{!}{
	\begin{tabular}{l|ccc}  
		\toprule  
		\multicolumn{4}{c}{\textbf{Column Prediction}} \\  
		\midrule  
		Method & \centering EM & Filtered \newline Acc & Rec \\  
		\midrule  
		Qwen2.5-0.5B (DTS-SQL) & 19.32 & 29.98 &  48.64  \\  
		Qwen2.5-1.5B (DTS-SQL) & 31.17 & 42.43 &  59.41  \\  \midrule
		Qwen2.5-0.5B (Cold-start) & 4.8 & 15.14 & 31.71  \\    
		Qwen2.5-1.5B (Cold-start)  & 19.42 & 35.05 &  54.52  \\  \midrule 
		Qwen2.5-0.5B (Schema-R1) & 13.24 & 44.02 & 64.86  \\  
		Qwen2.5-1.5B (Schema-R1) & \textbf{38.24} & \textbf{68.82} & \textbf{81.85}  \\  
		\bottomrule  
	\end{tabular} }
    \label{box2}
    \vspace{-12pt}
\end{table*} 

\textbf{\textit{RL training exhibiting superior results compared to SFT.}}
Tables~\ref{box1} and~\ref{box2} present a comparative analysis between supervised fine-tuning (SFT) and reasoning RL on our validation set. In the single-dataset configuration, the 0.5B model achieved a higher filtered accuracy than the DTS-SQL  (which relies solely on SFT) for both table and column prediction. In particular, the 1.5B model consistently outperformed DTS-SQL in all evaluation metrics, with particularly significant improvements across all metrics for both table and column prediction. These results suggest that larger models benefit more effectively from reasoning training, highlighting the scalability advantages of our approach.

\textbf{\textit{Reasoning-intensive tasks benefit more from Reasoning-Table.}}
The experimental results reveal an important finding: while training with limited high-quality CoT samples underperforms comprehensive supervised fine-tuning (as shown in comparison with Cold-start and DTS-SQL), a two-stage approach proves remarkably effective. Initial fine-tuning of the small sample set followed by GRPO-based reasoning training enables the model to autonomously refine its reasoning capabilities. This methodology allows the model to not only compensate for the limitations of small-sample training, but ultimately exceed the performance of conventional SFT approaches by internalizing and optimizing the acquired reasoning patterns.

\begin{figure}[t!]
    \centering
    \vspace{-18pt}
    \includegraphics[width=1.0\linewidth]{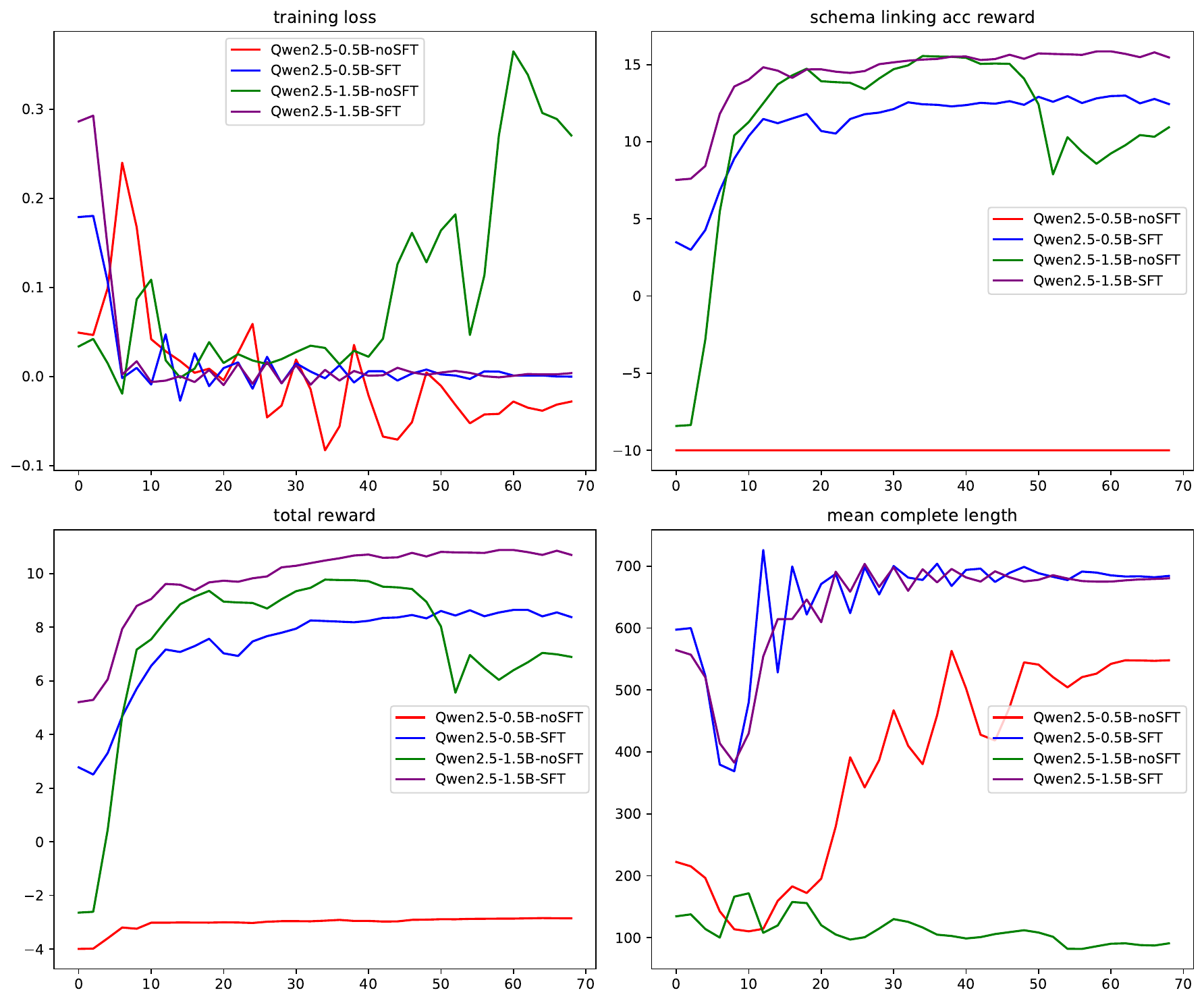} 
    \caption{The training process of schema-R1 on different base models}
    \label{fig:pic2} 
    \vspace{-18pt}
\end{figure}

\textbf{\textit{Training progress shows consistent improvement in reward and performance.}}
Fig.~\ref{fig:pic2} shows the progression of reward scores, response lengths, and evaluation metrics in training iterations. Notably, the cold-start SFTed model shows steady reward score improvement during training, contrasting sharply with the no-SFT baseline's poor performance and frequent instruction-following failures,especially in the 0.5B model. This compelling evidence suggests that precise intent recognition enables reinforcement learning to systematically optimize both the quality of model outputs and the effectiveness of task execution.

\section{Conclusion}
In this paper, we address the critical challenge of schema linking in Text-to-SQL systems by proposing Schema-R1, a robust reasoning-enhanced model that significantly improves accuracy and adaptability. By decoupling schema linking from SQL generation and enhancing it with self-improving reasoning, Schema-R1 paves the way for more accurate, adaptable, and privacy-conscious database interaction systems.
 
\noindent \textbf{Limitations}
Although our Schema-R1 has shown success in smaller models, computational constraints have prevented testing in larger models. Future work will involve validating Schema-R1 at a larger scale model.

Text-to-SQL, as a natural language-to-SQL task, presents a challenge: even with schema linking decoupled, aligning its predictions with the final SQL remains non-trivial. To address this, we will investigate self-correction models to ensure coherence between schema linking and SQL generation.

\clearpage 
\bibliographystyle{unsrt}  
\bibliography{references}

\clearpage 
\appendix
\section{Prompt Construction}
\label{app:slice-construction}
\begin{figure}[htbp] 
    \centering
    \includegraphics[width=1.0\linewidth]{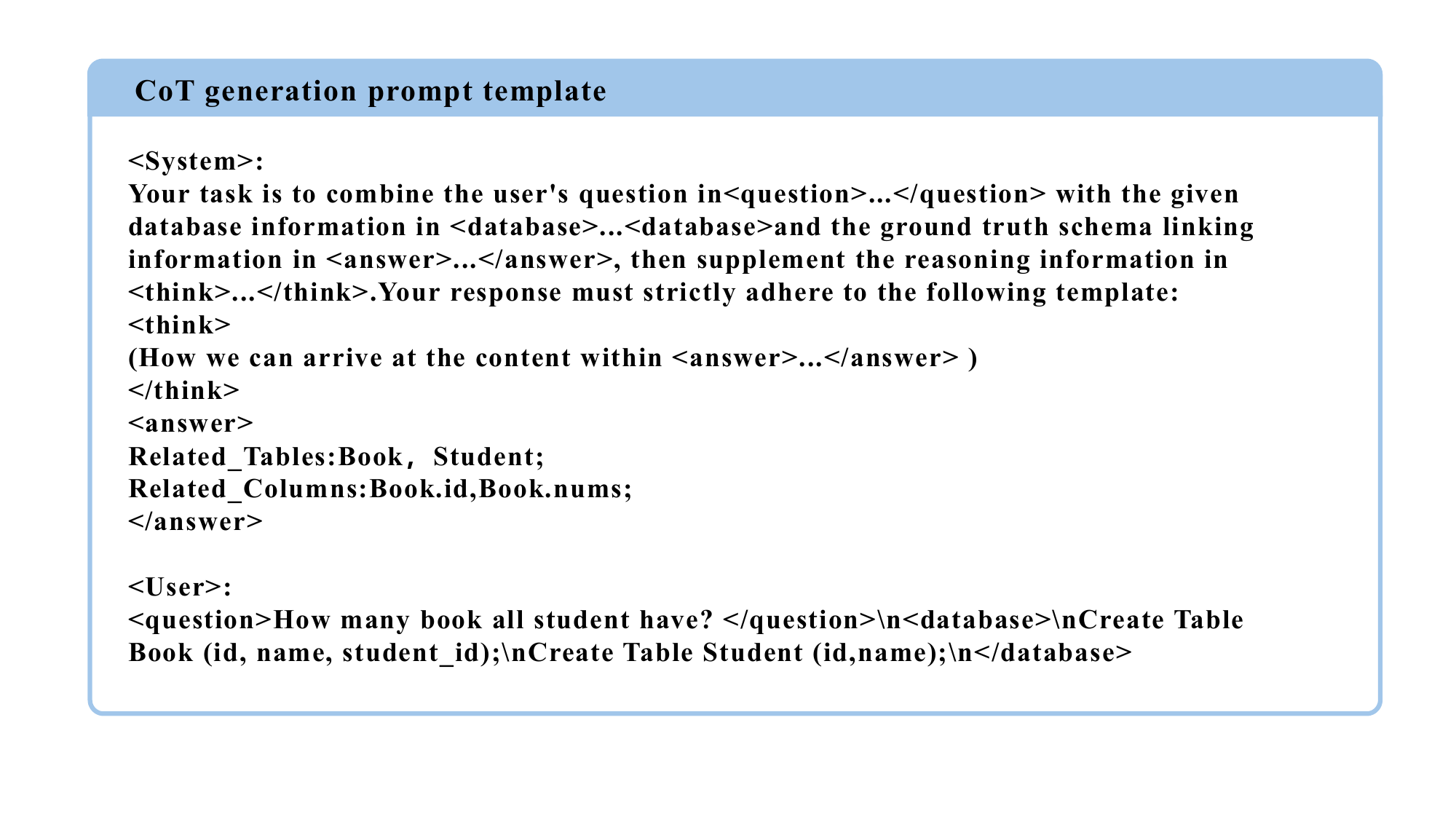}
    \vspace{-8pt}
    \caption{Prompt template for CoT generation}
    \vspace{-15pt}
\end{figure}

\end{document}